\pdfoutput=1

\documentclass[11pt]{article}

\usepackage[final]{acl}

\usepackage{times}
\usepackage{latexsym}

\usepackage[T1]{fontenc}

\usepackage[utf8]{inputenc}

\usepackage{microtype}

\usepackage{inconsolata}

\usepackage{graphicx}
\usepackage{multicol}
\usepackage{multirow}
\usepackage{amsmath}
\usepackage{amsfonts}
\usepackage{graphicx}
\usepackage{booktabs}
\usepackage{subcaption}
\usepackage{enumitem}
\usepackage{url}

\usepackage[normalem]{ulem}

\title{GMU Systems for the IWSLT 2025 Low-Resource Speech Translation Shared Task}

\author{Chutong Meng and Antonios Anastasopoulos \\
 George Mason University \\
  \texttt{\{cmeng2,antonis\}@gmu.edu}}

\begin{document}
\maketitle
\begin{abstract}
This paper describes the GMU systems for the IWSLT 2025 low-resource speech translation shared task.
We trained systems for all language pairs, except for Levantine Arabic.
We fine-tuned SeamlessM4T-v2 \cite{2023seamless} for automatic speech recognition (ASR), machine translation (MT), and end-to-end speech translation (E2E ST).
The ASR and MT models are also used to form cascaded ST systems.
Additionally, we explored various training paradigms for E2E ST fine-tuning, including direct E2E fine-tuning, multi-task training, and parameter initialization using components from fine-tuned ASR and/or MT models.
Our results show that
(1) direct E2E fine-tuning yields strong results;
(2) initializing with a fine-tuned ASR encoder improves ST performance on languages SeamlessM4T-v2 has not been trained on;
(3) multi-task training can be slightly helpful.\footnote{We release our code for reproducibility: \url{https://github.com/mct10/IWSLT2025_LowRes_ST}.}
\end{abstract}

\section{Introduction}

\begin{figure}[t]
    \centering
    \includegraphics[width=0.90\linewidth]{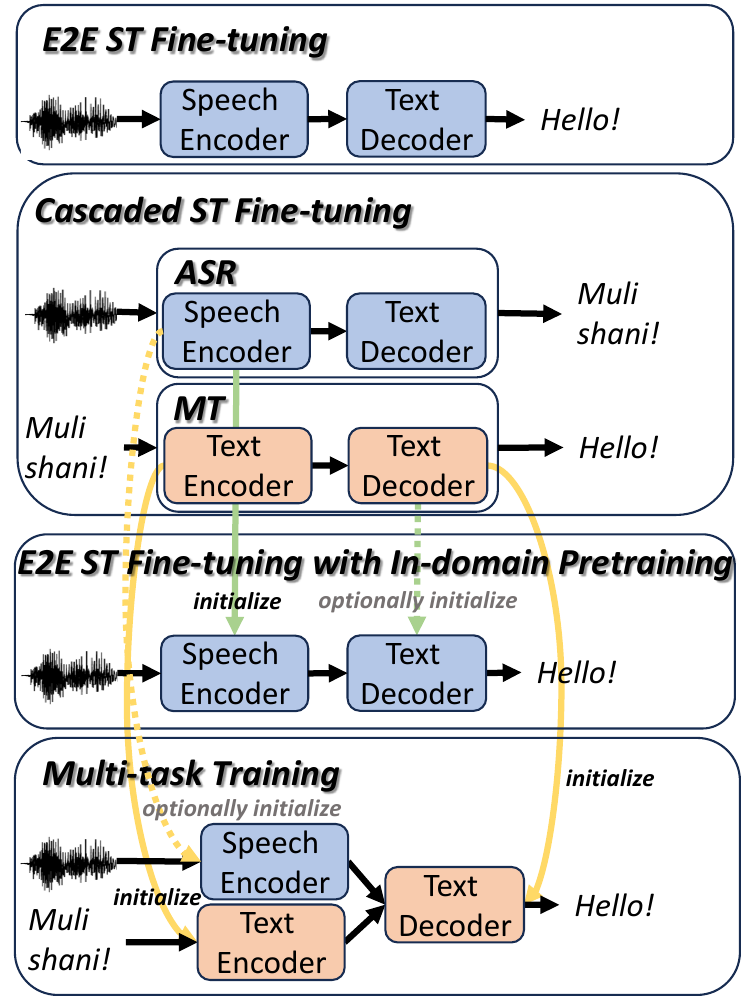}
    \caption{Illustration of our SeamlessM4T-v2 fine-tuning strategies.
    Speech Encoder, Text Encoder, and Text Decoder refer to the corresponding components of SeamlessM4T-v2. 
    }
    \label{fig:figure1}
\vspace{-10pt}
\end{figure}

Speech translation (ST) is a task that aims to translate speech in one language into text in another language.
It can be addressed by either an end-to-end (E2E) ST model or a cascaded system that combines an automatic speech recognition (ASR) model and a machine translation (MT) model.
Recent advances in E2E ST have been driven by the development of large multilingual models trained on large amounts of multilingual datasets \cite{2023seamlessm4t, 2023seamless, Radford-2023-whisper}.
Similar trends can be observed in ASR~\cite{Radford-2023-whisper} and~MT \cite{nllb2022} as well.
Despite these models have covered a wide range of languages, many low-resource languages remain underrepresented and are not yet well supported by existing models.

The IWSLT low-resource speech translation shared tasks~\cite{abdulmumin-etal-2025-findings,ahmad-etal-2024-findings,agrawal-etal-2023-findings,anastasopoulos-etal-2022-findings} are designed to advance ST technology for low-resource languages. 
To address the challenge of data scarcity, previous submissions have explored various pre-trained models, including multilingual self-supervised speech models such as XLSR \cite{conneau21xlsr}, multilingual ASR models such as Whisper \cite{Radford-2023-whisper}, multilingual MT models such as NLLB~\cite{nllb2022}, and multilingual ST models such as SeamlessM4T~\cite{2023seamlessm4t,2023seamless}.
These pre-trained models were then fine-tuned on ST datasets for low-resource languages.
Among them, SeamlessM4T-v2 has demonstrated superior performance, according to last year's evaluations~\cite{ahmad-etal-2024-findings}.

This paper describes GMU submissions to the IWSLT 2025 low-resource speech translation task~\cite{abdulmumin-etal-2025-findings}.
Our work focuses on fine-tuning the SeamlessM4T-v2 model \cite{2023seamless} for all language pairs except Levantine Arabic-to-English.
We fine-tuned the model for both E2E and cascaded systems.
For E2E ST fine-tuning, we explored multiple strategies, including multi-task training with MT and knowledge distillation objectives, as well as initializing model components with those from fine-tuned ASR and/or MT models, trying to utilize all available datasets.
Figure~\ref{fig:figure1} illustrates our strategies.
Our results show that direct E2E fine-tuning SeamlessM4T-v2 yields strong performance across all languages pairs, except Quechua, which has too little training data.
For languages not seen during SeamlessM4T-v2 pre-training, we show that fine-tuning the model on ASR data and initializing the ST encoder with the ASR encoder improves performance significantly.
We also show that multi-task training offers some performance gains when the MT model significantly outperforms the E2E ST model.

\begin{table*}[t]
    \footnotesize
    \centering
    \begin{tabular}{cccc}
        \toprule
        \textbf{Language} & \textbf{Task} & \textbf{Amount} & \textbf{Sources} \\
        \midrule
        \multirow{2}{*}{aeb-eng} & ASR & 156 hours & \multirow{2}{*}{LDC2022E01} \\
         & 3-way ST & 161 hours/202k lines & \\
         \midrule
         bem-eng & 3-way ST & 167 hours/82k lines & \citet{sikasote-etal-2023-big} \\
         \midrule
         fon-fra & 2-way ST & 47 hours & IWSLT2025 \cite{abdulmumin-etal-2025-findings} \\
         \midrule
         \multirow{3}{*}{gle-eng} & ASR & 5 hours & CommonVoice 21.0 \\
         & 2-way ST & 7 hours & IWSLT2025 \\
         & 3-way ST & 202 hours &  \citet{moslem2024leveraging} \\
         \midrule
         \multirow{2}{*}{bho-hin} & 2-way ST & 20 hours & IWSLT2025 \\
         & ASR & 60 hours & ULCA \\
         \midrule
         est-eng & 3-way ST & 1213 hours/581k lines & IWSLT2025 \\
         \midrule
         mlt-eng & 3-way ST & 12 hours/9k lines & IWST2025 \\
         \midrule
         \multirow{2}{*}{mar-hin} & ASR & 15 hours & CommonVoice 21.0; \citet{he-etal-2020-open} \\
         & 2-way ST & 16 hours & IWSLT2025 \\
         \midrule
         \multirow{3}{*}{que-spa} &  MT & 46k lines & \citet{ortega2020neural}; \citet{nllb2022} \\
         & ASR & 48 hours & \citet{cardenas2018siminchik}\\
         & 3-way ST & 9 hours/2k lines & IWSLT2025; \citet{zevallos2022huqariq} \\
         \bottomrule
    \end{tabular}
    \caption{Summary of datasets used for training.
    2-way ST refers to datasets with paired source speech and target text, while 3-way ST includes paired source speech, source text, and target text.
    The 3-way ST datasets can be used for ASR and MT training as well.}
    \label{tab:data}
\end{table*}

\section{Task Descriptions}
The IWSLT 2025 low-resource ST task~\cite{abdulmumin-etal-2025-findings} covers 10 language pairs: (North) Levantine Dialectal Arabic to English (\verb|apc|-\verb|eng|), Tunisian Arabic Dialect to English (\verb|aeb|-\verb|eng|), Bemba to English (\verb|bem|-\verb|eng|), Fongbe to French (\verb|fon|-\verb|fra|), Irish to English (\verb|gle|-\verb|eng|), Bhojpuri to Hindi (\verb|bho|-\verb|hin|), Estonian to English (\verb|est|-\verb|eng|), Maltese to English (\verb|mlt|-\verb|eng|), Marathi to Hindi (\verb|mar|-\verb|hin|), and Quechua to Spanish (\verb|que|-\verb|spa|).
In each of these language pairs, the source language is low-resource while the target language is high-resource.
We trained systems for all language pairs except for \verb|apc|-\verb|eng|.\footnote{The LDC resources for apc cannot be obtained for free this year.} 

Formally, E2E ST is defined as translating a speech utterance $x^{\text{sp}}$ in the source language into text $y$ in the target language.
For cascaded ST, a source speech utterance $x^{\text{sp}}$ is first transcribed into text $x^{\text{text}}$ in the source language using an ASR model, which is then translated into the target-language text $y$ using an MT model.

The datasets we used are summarized in Table \ref{tab:data}.
Each of the official datasets provided by the organizers is either a 2-way ST or a 3-way ST dataset.
A 2-way ST data sample is represented as a tuple $(x^{\text{sp}}, y)$, while a 3-way ST data sample refers to a triple $(x^{\text{sp}}, x^{\text{text}}, y)$.
3-way ST datasets are available for \verb|aeb-eng|, \verb|bem-eng|, \verb|est-eng|, \verb|mlt-eng|, and \verb|que-spa|.
The other languages are provided with 2-way ST datasets.
Among these, \verb|est-eng| has the largest dataset with more than 1,000 hours of speech.
Both \verb|aeb-eng| and \verb|bem-eng| have more than 100 hours of data, while datasets for other languages are limited and having only about 10 hours of speech.
In addition, the organizers provide pointers to additional ASR and MT datasets.
An ASR data sample is represented as $(x^{\text{sp}}, x^{\text{text}})$, while an MT data sample is represented as $(x^{\text{text}}, y)$.
It is evident that both ASR and MT datasets can be derived from 3-way ST datasets.

The task allows submissions under two conditions: constrained and unconstrained.
Under the constrained condition, only the provided dataset can be used and no pre-trained models are allowed. 
The unconstrained condition allows the use of any models and any datasets.
All of our submissions fall under the unconstrained condition.

\section{Methods}
\label{sec:methods}
Our methods focus on fine-tuning the SeamlessM4T-v2 model \cite{2023seamless}.
We explore 4 different fine-tuning strategies:
(1) E2E ST fine-tuning; 
(2) ASR and MT fine-tuning for the cascaded system; 
(3) multi-task training similar to \citet{2023seamless}; 
(4) initializing ST model components with those from ASR and/or MT models.
We fine-tune the model on a single language pair at a time.
Due to the dataset availability and model performance for each language pair, not all strategies have been tried for every pair.

Although the MT components of SeamlessM4T-v2 are initialized by the NLLB model \cite{nllb2022}, SeamlessM4T-v2 has been trained on less languages and supports MT for only 4 out of the 10 language pairs in this shared task.
In contrast, the NLLB model supports MT for all 10 pairs.
To evaluate whether the smaller language coverage of SeamlessM4T-v2 impacts performance, we additionally fine-tuned an NLLB model on the MT datasets, using it as the MT baseline.
Section~\ref{sec:model} introduces the NLLB and SeamlessM4T-v2 models.
Section \ref{sec:e2e} through Section \ref{sec:pretrain} elaborate our fine-tuning strategies.

\subsection{Base Models}
\label{sec:model}
\textbf{NLLB} \cite{nllb2022}. 
NLLB is a multilingual MT model supporting over 200 languages, including all language pairs in this shared task.
The model is available in two architecture variants: a sparsely gated mixture-of-experts (MoE) one and a set of dense transformer models.
The dense transformer architecture comprises a text encoder and a text decoder.
While the MoE variant (NLLB-200) achieves the strongest performance, it has 54.5B parameters and is not practical for fine-tuning.
Therefore, in our experiments, we choose the 1.3B dense transformer model distilled from NLLB-200, referred to as \verb|NLLB-200-Distilled-1.3B|.

\textbf{SeamlessM4T-v2} \cite{2023seamless}.
SeamlessM4T-v2 is the state-of-the-art foundation model for ST.
While it supports into-speech translation, we only focus on its into-text translation capabilities for the purpose of this shared task.
SeamlessM4T-v2 is composed of a speech encoder, a text encoder, and a shared text decoder.
Its \verb|Large| variant has 2B parameters in total and we refer to it as \verb|SeamlessM4T-v2-Large|.
The speech encoder is pre-trained on 4.5M hours of unlabeled audio with the w2v-BERT 2.0 objective.
The text encoder and decoder are initialized by the NLLB model.
During fine-tuning, a multi-task training strategy is employed, incorporating ASR, MT, ST, and knowledge distillation (KD) objectives.
We also explore this strategy in our experiments.
The model supports 101 languages for speech input and 96 languages for text input and output.
Among the low-resource languages in this shared task, SeamlessM4T-v2 supports \verb|est|, \verb|gle|, \verb|mar|, and \verb|mlt|, but does not support \verb|aeb|, \verb|bem|, \verb|bho|, \verb|fon|, and \verb|que|.

\subsection{E2E ST Fine-tuning}
\label{sec:e2e}
For E2E fine-tuning, we utilize 2-way ST data samples $(x^{\text{sp}}, y)$.
We use Equation \ref{eq:e2e} as the loss function to optimize the speech encoder and the text decoder.
\begin{align}
\label{eq:e2e}
\begin{split}
L_{\text{E2E}}=&
-\frac{1}{|y|}\log p(y|x^{\text{sp}};\theta_{\text{se}},\theta_{\text{td}}) \\
&=-\frac{1}{|y|}\sum_{i=1}^{|y|}\log p(y_i|y_{<i},x^{\text{sp}};\theta_{\text{se}},\theta_{\text{td}})
\end{split}
\end{align}
$\theta_{\text{se}}$ and $\theta_{\text{td}}$ denote the parameters of the speech encoder and the text decoder, respectively.

\subsection{ASR and MT Fine-tuning for the Cascaded ST System}
\label{sec:cascade}
Since SeamlessM4T-v2 also supports multilingual ASR and MT, it is suitable for being fine-tuned on the low-resource languages for ASR and MT as well.
Specifically, ASR data samples $(x^{\text{sp}},x^{\text{text}})$ and MT data samples $(x^{\text{text}}, y)$ are used.
A cascaded system can then be built by a fine-tuned ASR  and a fine-tuned MT model.
The corresponding loss functions for ASR and MT fine-tuning are defined in Equation \ref{eq:asr} and Equation \ref{eq:mt}, respectively.
Equation \ref{eq:mt} is also used for NLLB MT fine-tuning.
\begin{align}
\small
\label{eq:asr}
\begin{split}
L_{\text{ASR}}=&
-\frac{1}{|x^{\text{text}}|}\log p(x^{\text{text}}|x^{\text{sp}};\theta_{\text{se}},\theta_{\text{td}}) \\
&=-\frac{1}{|x^{\text{text}}|}\sum_{i=1}^{|x^{\text{text}}|}\log p(x^{\text{text}}_i|x^{\text{text}}_{<i},x^{\text{sp}};\theta_{\text{se}},\theta_{\text{td}})
\end{split}
\end{align}
\begin{align}
\label{eq:mt}
\begin{split}
L_{\text{MT}}=&
-\frac{1}{|y|}\log p(y|x^{\text{text}};\theta_{\text{te}},\theta_{\text{td}}) \\
&=-\frac{1}{y}\sum_{i=1}^{|y|}\log p(y_i|y_{<i},x^{\text{text}};\theta_{\text{te}},\theta_{\text{td}})
\end{split}
\end{align}
$\theta_{\text{te}}$ refers to the parameters of the text encoder.
We use $\theta_{\text{se}}^{\text{ASR}}$ and $\theta_{\text{td}}^{\text{ASR}}$ to denote the fine-tuned ASR components, $\theta_{\text{te}}^{\text{MT}}$ and $\theta_{\text{td}}^{\text{MT}}$ to denote the fine-tuned MT components.

\subsection{Multi-task Fine-tuning}
\label{sec:mlt}
Inspired by the multi-task fine-tuning strategy in \citet{2023seamless}, we adopt a similar approach and explore its effect in the low-resource ST setting.

Our approach includes ST, MT, and KD objectives, using paired 3-way ST data samples $(x^{\text{sp}}, x^{\text{text}}, y)$.
The ST objective follows Equation~\ref{eq:e2e} and the MT objective follows Equation \ref{eq:mt}.
The goal of applying the KD objective is to use the MT components to enhance the ST components.
The motivation is that MT is generally an easier task than ST and often yields better performance, and we hope to mitigate this performance gap.
In order to have a strong MT teacher, we initialize the text encoder and the text decoder in SeamlessM4T-v2 with $\theta_{\text{te}}^{\text{MT}}$ and $\theta_{\text{td}}^{\text{MT}}$ from Section \ref{sec:cascade}, respectively.
Optionally, to help with convergence, we can initialize the speech encoder with $\theta_{\text{se}}^{\text{ASR}}$.
Equation~\ref{eq:kd_teacher} explains how we obtain the teacher probability distribution from the MT components. 
\begin{align}
\label{eq:kd_teacher}
\begin{split}
&p_{\text{teacher}}(\cdot|y_{<i}, x^{\text{text}}) \\
&=\text{stop-gradient}\left(p(\cdot|y_{<i}, x^{\text{text}};\theta_{\text{te}},\theta_{\text{td}})\right)
\end{split}
\end{align}
$\text{stop-gradient}(\cdot)$ means that we detach the resultant tensor from the computation graph, thereby preventing the gradients from the teacher probability distribution being propagated to the MT teacher parameters $\theta_{\text{te}}$ and $\theta_{\text{td}}$.
We tried without $\text{stop-gradient}(\cdot)$ but observed a performance drop.
Then, we compute KL-Divergence between the student and the teacher probability distributions with Equation \ref{eq:kd}.
\begin{align}
\label{eq:kd}
\footnotesize
\begin{split}
&L_{\text{KD}}\\
&=\frac{1}{|y|}\sum_{i=1}^{|y|}D_{\text{KL}}\left[p_{\text{teacher}}(\cdot|y_{<i}, x^{\text{text}}) || p(\cdot|y_{<i}, x^{\text{sp}}; \theta_{\text{se}},\theta_{\text{td}})\right] \\
&=\frac{1}{|y|}\sum_{i=1}^{|y|}\left[ p_{\text{teacher}}(\cdot|y_{<i}, x^{\text{text}}) \cdot \log \frac{ p_{\text{teacher}}(\cdot|y_{<i}, x^{\text{text}})}{p(\cdot|y_{<i}, x^{\text{sp}}; \theta_{\text{se}},\theta_{\text{td}})}\right]
\end{split}
\end{align}
The student probability distribution comes from the ST components $\theta_{\text{se}}$ and $\theta_{\text{td}}$.

During fine-tuning, $\theta_{\text{se}}$ and $\theta_{\text{td}}$ are updated while $\theta_{\text{te}}$ is kept frozen.
The final loss function is a linear combination of the three losses:
\begin{align}
L=\alpha\cdot L_{\text{E2E}} + \beta \cdot L_{\text{MT}} + \gamma \cdot L_{\text{KD}}
\end{align}
where $\alpha$, $\beta$, and $\gamma$ are constants which can be tuned on the development set.
Empirically, we found that $\alpha=1$, $\beta=1$, and $\gamma=2$ worked best.

\subsection{E2E ST Fine-tuning with In-domain Pre-trained Components}
\label{sec:pretrain}
As mentioned in Section~\ref{sec:model}, the SeamlessM4T-v2 model has not been trained on 5 low-resource languages of interest.
To better adapt the model to new languages for ST, we fine-tune it on in-domain ASR data with Equation~\ref{eq:asr}, such that the fine-tuned speech encoder $\theta_{\text{sp}}^{\text{ASR}}$ can better capture semantics from the speech in the new language.
Then, we can initialize the speech encoder of the ST model with $\theta_{\text{sp}}^{\text{ASR}}$ for E2E ST fine-tuning.
Optionally, we can also initialize the text decoder by $\theta_{\text{td}}^{\text{MT}}$.
However, we do not expect the fine-tuned decoder to be as helpful as the fine-tuned speech encoder, as the target language is always high-resource and the SeamlessM4T-v2 model has been trained on a lot of that.
After the initializations, we perform E2E ST fine-tuning with Equation \ref{eq:e2e}.

\section{Experiments}
We describe the additional datasets we used in Section \ref{sec:dataset}.
In Section \ref{sec:exp_setup}, we describe the fine-tuning hyperparameters.
The evaluation metrics are described in Section \ref{sec:metrics}.

\subsection{Dataset}
\label{sec:dataset}
All datasets are summarized in Table \ref{tab:data}.
Besides the official ST datasets provided by the organizers, we use the following additional datasets.

\textbf{gle-eng.}
We use the synthetic 3-way ST dataset from \citet{moslem2024leveraging}.
The text is extracted from OPUS \cite{tiedemann-2012-opus}, covering portions of the Wikimedia, Tatoeba, and EUbbookshop corpora.
The speech is synthesized using the Azure Speech service.
This synthetic dataset has about 202 hours of speech.
We also use the \verb|gle| ASR dataset from CommonVoice 21.0\footnote{\url{https://commonvoice.mozilla.org/en/datasets}}~\cite{ardila-etal-2020-common} to include real speech data.

\textbf{bho-hin.}
We use the \verb|bho| dataset from the ULCA corpus.\footnote{\url{https://github.com/Open-Speech-EkStep/ULCA-asr-dataset-corpus}}
It has 60 hours of speech.

\textbf{mar-hin.}
We collect Marathi ASR data from CommonVoice \cite{ardila-etal-2020-common} and OpenSLR64 \cite{he-etal-2020-open}, totaling 15 hours of speech.

\textbf{que-spa.}
The official 3-way ST dataset has merely 1.6 hours of speech, so we try to find and use as much data as possible.
We use the additional synthetic 3-way ST dataset \cite{zevallos2022huqariq}, whose Spanish translations are generated by Google Translate.
We also include the additional 48-hour ASR dataset \cite{cardenas2018siminchik}.
For MT, we use the additional MT dataset \cite{ortega2020neural} extracted from JW300 and Hinantin.
The data is very noisy, so we apply extensive text cleaning strategies inspired by \citet{koehn-etal-2018-findings}.
Furthermore, we obtain the NLLB Quechua-English dataset from OPUS\footnote{\url{https://opus.nlpl.eu/NLLB/qu&en/v1/NLLB}} \cite{tiedemann-2012-opus}.
This dataset is obtained by text mining~\cite{fan2020beyond, schwenk-etal-2021-ccmatrix}.
We translate the English text into Spanish by applying \verb|NLLB-200-Distilled-1.3B|, creating a synthetic Quechua-Spanish MT dataset having approximately 34k lines.

In general, for ASR, MT, and E2E ST experiments, we use their designated datasets as well as subsets extracted from the 3-way ST datasets if available.

In our experiments, we keep the text in their original form.
No text normalization is performed, except for apostrophe normalization in \verb|fon|.
All speech files are resampled to 16khz if they originally have a different sampling rate.

\subsection{Experiment Setup}
\label{sec:exp_setup}
We fine-tune \verb|SeamlessM4T-v2-Large| for all language pairs.
Two codebases are used in our experiments. 
One is the official repository,\footnote{\url{https://github.com/facebookresearch/seamless_communication}} which is for E2E ST fine-tuning (Section \ref{sec:e2e}) only.
To support all fine-tuning strategies, we have implemented a second codebase based on the HuggingFace Transformers toolkit\footnote{\url{https://huggingface.co/docs/transformers/main/model_doc/seamless_m4t_v2}} \cite{wolf-etal-2020-transformers}.
The HuggingFace codebase is designed to be identical to the official fine-tuning script. 
However, in practice, we observed some performance gaps between the two codebases, which we discuss in detail in Appendix~\ref{app:codebase}.

Additionally, we fine-tune \texttt{NLLB-200-} \texttt{Distilled-1.3B} as the MT baseline (the reason is discussed in Section~\ref{sec:methods}).

For all experiments, we use the AdamW optimizer \cite{loshchilov2019adamw} with betas~$(0.9, 0.98)$, and no weight decay.
Models are trained for a maximum of~10 epochs.
We use a learning rate of~1e-4, with the first epoch being the warmup phase.
For E2E ST fine-tuning with model components initialized by ASR and/or MT components, we use a smaller learning rate of~6e-5.
We use the inverse square root learning rate scheduler.
The batch size is~120 utterances for speech input tasks (ASR and ST) and~256 sentences for text input tasks (MT).
The label smoothing weight is~0.2.
These hyperparameters can be slightly adjusted for different language pairs depending on dataset characteristics.
For instance, for \verb|que-spa|, we use a learning rate of~1e-5 for ST fine-tuning and the maximum epoch number is~200.
For ASR fine-tuning on \verb|est-eng| and \verb|que-spa|, the batch size is~72 utterances due to longer input durations.
Lastly, for MT fine-tuning, the hyperparameters for NLLB are exactly the same as \texttt{SeamlessM4T-v2}.
During inference, we use a beam size of 5 and length penalty of 1.0.

\subsection{Evaluation Metrics}
\label{sec:metrics}
We evaluate ASR performance using word error rate (WER) and character error rate (CER) with the \verb|jiwer|\footnote{\url{https://github.com/jitsi/jiwer}} package.
For MT performance, we use SacreBLEU\footnote{\url{https://github.com/mjpost/sacrebleu}} \cite{post-2018-call} to compute BLEU\footnote{Signature: nrefs:1 + case:lc + eff:no + tok:13a + smooth:exp + version:2.5.1} scores.
For both evaluations, text is lowercased and punctuations are removed before scoring.

\section{Results and Analysis}
We first present the ASR and MT performance in Section \ref{sec:asr_res} and \ref{sec:mt_res}, respectively.
Then, we summarize the ST performance in Section \ref{sec:st_res}.
The ablation study of using additional datasets is presented in Appendix \ref{app:ablate}.

\subsection{Automatic Speech Recognition}
\label{sec:asr_res}

\begin{table}[t]
    \small
    \centering
    \begin{tabular}{cccccc}
    \toprule
        \multirow{2}{*}{\textbf{Lang}} & \multirow{2}{*}{\textbf{System}} & \multicolumn{2}{c}{\textbf{Dev}} & \multicolumn{2}{c}{\textbf{Public Test}} \\
        \cmidrule{3-4} \cmidrule{5-6}
        & & CER & WER & CER & WER \\
        \midrule
        aeb & Seamless-FT & 20.7 & 41.2 & 24.6 & 49.0 \\
        \midrule
        bem & Seamless-FT & 9.27 & 31.08 & 8.86 & 30.40 \\
        \midrule
        \multirow{2}{*}{gle} & Seamless-0s & 14.27 & 23.90 & 14.79 & 24.61 \\
        & Seamless-FT & 5.51 & 9.47 & 4.71 & 8.39 \\
        \midrule
        bho$^\dagger$ & Seamless-FT & 32.68 & 41.86 & - & - \\
        \midrule
        \multirow{2}{*}{est} & Seamless-0s & 12.94 & 22.22 & - & - \\
        & Seamless-FT & 3.06 & 8.59 & - & - \\
        \midrule
        \multirow{2}{*}{mlt} & Seamless-0s & 8.57 & 20.68 & - & - \\
        & Seamless-FT & 3.69 & 12.12 & - & - \\
        \midrule
        \multirow{2}{*}{mar$^\dagger$} & Seamless-0s & 4.28 & 17.40 & 4.73 & 18.44 \\
        & Seamless-FT & 1.90 & 8.42 & 8.15 & 2.08 \\
        \midrule
        que & Seamless-FT & 15.54 & 37.80 & - & - \\
        \bottomrule
    \end{tabular}
    \caption{ASR results for languages with available ASR datasets.
    $^\dagger$: Models are \textbf{not} evaluated on official IWSLT2025 datasets but on additional ASR datasets.
    The bho model is evaluated on ULCA, and the mar model is evaluated on CommonVoice.
    \textbf{0s} denotes a zero-shot model, while \textbf{FT} denotes a fine-tuned model.
    }
    \label{tab:asr_res}
\end{table}

\begin{table}[t]
    \small
    \centering
    \begin{tabular}{cccccc}
    \toprule
        \multirow{2}{*}{\textbf{Lang}} & \multirow{2}{*}{\textbf{System}} & \multicolumn{2}{c}{\textbf{Eval 1}} & \multicolumn{2}{c}{\textbf{Eval 2}} \\
        \cmidrule{3-4} \cmidrule{5-6}
        & & CER & WER & CER & WER \\
        \midrule
        aeb & Seamless-FT & 19.7 & 38 & 22.3 & 39.9 \\
        \midrule
        bem & Seamless-FT & 8.96 & 30.62 & - & - \\
        \bottomrule
    \end{tabular}
    \caption{Official ASR Evaluation results for aeb and bem.
    We did not submit hypothesis for other language pairs unfortunately.
    }
    \label{tab:asr_res_off}
\vspace{-10pt}
\end{table}

Internal evaluation results are presented in Table~\ref{tab:asr_res}, and the official evaluation results~\cite{abdulmumin-etal-2025-findings} are in Table~\ref{tab:asr_res_off}.
\textbf{Seamless-FT} refers to \verb|SeamlessM4T-v2-Large| fine-tuned on all available ASR datasets, while \textbf{Seamless-0s} refers to \verb|SeamlessM4T-v2-Large| evaluated in a zero-shot manner without fine-tuning.
Languages without zero-shot results are not supported by SeamlessM4T-v2's ASR capability.
For \verb|bho| and \verb|mar|, no official ASR datasets are provided by the organizers, so we evaluate them on held-out subsets from ULCA and CommonVoice, respectively.

The zero-shot performances on \verb|gle|, \verb|est|, \verb|mlt|, and \verb|mar| are relatively strong, all with WER around 20\% and CER around 10\%.
Further fine-tuning on in-domain ASR datasets yields substantial improvements, reducing both CER and WER by about 50\% in relative value.
For languages that SeamlessM4T-v2 has not been trained on, ASR performance is poorer.
\verb|aeb| and \verb|bho| are particularly challenging, with WERs greater than 40\%.
\verb|que| also has a high WER of 37.8\%.
The model performs relatively well on \verb|bem|, achieving a low CER of approximately 9\%, although its WER remains high at around 30\%.
We can conclude from these results that the fine-tuned \verb|SeamlessM4T-v2-Large| performs better on languages it has been trained on.

\subsection{Text Machine Translation}
\label{sec:mt_res}

\begin{table}[t]
    \small
    \centering
    \begin{tabular}{cccc}
    \toprule
        \multirow{2}{*}{\textbf{Lang}} & \multirow{2}{*}{\textbf{System}} & \multicolumn{1}{c}{\textbf{Dev}} & \multicolumn{1}{c}{\textbf{Public Test}} \\
        & & BLEU & BLEU \\
        \midrule
        \multirow{3}{*}{aeb} & NLLB-0s & 11.05 & 8.98  \\
        & NLLB-FT & \textbf{30.48} & 27.11  \\
        & Seamless-FT & 30.39 & \textbf{27.54} \\
        \midrule
        \multirow{3}{*}{bem} & NLLB-0s & 8.57 & 8.58 \\
        & NLLB-FT & \textbf{29.20} & \textbf{30.42} \\
        & Seamless-FT & 28.86 & 29.27 \\
        \midrule
        \multirow{4}{*}{est} & NLLB-0s & 31.60 & - \\
        & Seamless-0s & 30.33 & - \\
        & NLLB-FT & 32.85 & - \\
        & Seamless-FT & \textbf{40.23} & - \\
        \midrule
        \multirow{4}{*}{mlt} & NLLB-0s & 50.39 & - \\
        & Seamless-0s & 53.96  & - \\
        & NLLB-FT & \textbf{64.29} & - \\
        & Seamless-FT & 62.13 & \\
        \midrule
        \multirow{3}{*}{que} & NLLB-0s & 5.05 & - \\
        & NLLB-FT & \textbf{15.98} & - \\
        & Seamless-FT & 15.29 & - \\
        \bottomrule
    \end{tabular}
    \caption{MT results for languages with available MT datasets.
    \textbf{0s} denotes a zero-shot model, while \textbf{FT} denotes a fine-tuned model.
    There are only small gaps between NLLB-FT and Seamless-FT.
    }
    \label{tab:mt_res}
\vspace{-10pt}
\end{table}

Table \ref{tab:mt_res} presents MT performance for languages with available MT datasets.
We report both 0-shot and fine-tuned results for \verb|SeamlessM4T-v2-Large| and \verb|NLLB-200-Distilled-1.3B|.
\textbf{NLLB-0s} and \textbf{Seamless-0s} refer to the zero-shot performance, while \textbf{NLLB-FT} and \textbf{Seamless-FT} refer to the fine-tuned results.
Note that NLLB results are used only for reference.
The fine-tuned \verb|NLLB-200-Distilled-1.3B| are neither used for submissions nor for model initialization.

Fine-tuning on the in-domain MT datasets leads to substantial improvements.
\verb|NLLB-200-Distilled-1.3B| achieves +10 BLEU for all languages, except for \verb|est|, where the gain is +1.25 BLEU.
For \verb|aeb| and \verb|bem|, the improvements even reach approximately +20 BLEU.
Despite being trained on fewer languages, the fine-tuned \verb|SeamlessM4T-v2-Large| achieves performance comparable to that of the fine-tuned \verb|NLLB-200-Distilled-1.3B|.
This justifies our choice of adopting \verb|SeamlessM4T-v2-Large| as the MT model.

\subsection{Speech Translation}
\label{sec:st_res}

\begin{table}[t]
    \small
    \centering
    \begin{tabular}{ccc}
    \toprule
        \textbf{Lang} & \textbf{Dev} & \textbf{Public Test} \\
        \midrule
        aeb & 4.29 & 3.22 \\
        bem & 0.93 & 0.93 \\
        fon & 1.09 & - \\
        gle & 28.98 & 47.66 \\
        bho & 25.28 & - \\
        est & 26.21 & - \\
        mlt & 50.02 & - \\
        mar & 24.07 & 31.77 \\
        que & 1.47 & - \\
        \bottomrule
    \end{tabular}
    \caption{Zero-shot SeamlessM4T-v2-Large ST results for all languages.
    Results are obtained using the official codebase.
    }
    \label{tab:0shot_st}
\vspace{-10pt}
\end{table}

\begin{table*}[t]
    \footnotesize
    \centering
    \begin{tabular}{cllcccc}
    \toprule
        \textbf{Lang} & \multicolumn{1}{c}{\textbf{System}} & \multicolumn{1}{c}{\textbf{Submission}} & \textbf{Dev}  & \textbf{Public Test} & \textbf{Eval 1} & \textbf{Eval 2} \\
        \midrule
        \multirow{7}{*}{aeb} 
        & HF-E2E-ASR$_{\text{init}}$ & Primary & \textbf{25.48} & \textbf{21.41} & \textbf{20.30} & \textbf{17.8}\textsuperscript{\dag} \\
        & HF-MLT-ASR$_{\text{init}}$ & Contrastive 1 & 24.64 & 21.18 & 19.2 & 17.3 \\
        & HF-Cascaded & Contrastive 2 & 24.42 & 21.01 & 18.90 & 17.3 \\
        & HF-MLT & \multicolumn{1}{c}{-} & 24.23 & 20.33 & - & - \\
        & HF-E2E-ASR$_{\text{init}}$-MT$_{\text{init}}$ & \multicolumn{1}{c}{-} & 24.08 & 20.41 & - & - \\
        & OFF-E2E & \multicolumn{1}{c}{-} & 23.76 & 19.67 & - & - \\
        & HF-E2E & \multicolumn{1}{c}{-} & 22.73 & 18.35 & - & - \\
        \midrule
        \multirow{4}{*}{bem}
        & HF-E2E-ASR$_{\text{init}}$ & Primary & \textbf{31.96} & \textbf{32.12} & \textbf{31.7} & - \\
        & HF-E2E & Contrastive 1 & 31.14 & 30.93 & 30.6 & - \\
        & HF-Cascaded & Contrastive 2 & 28.02 & 28.02 & 27.9 & - \\
        & OFF-E2E & \multicolumn{1}{c}{-} & 30.69 & 31.23 & - & - \\
        \midrule
        fon & OFF-E2E & Primary & \textbf{40.86} & - & \textbf{31.96} & - \\
        \midrule
        \multirow{3}{*}{gle\textsuperscript{*}} 
        & OFF-E2E & Primary & \textbf{29.63} & 51.91 & \textbf{13.4} & - \\
        & HF-E2E & Contrastive 1 & 24.07 & 51.21 & 8.4 & - \\
        & HF-E2E-ASR$_{\text{init}}$ & Contrastive 2 & 23.34 & 51.43 & 6.7 & - \\
        \midrule
        \multirow{3}{*}{bho} 
        & OFF-E2E & Primary & \textbf{41.96} & - & \textbf{3.9} & - \\
        & HF-E2E-ASR$_{\text{init}}$ & Contrastive 1 & 39.04 & - & 3.4 & - \\
        & HF-E2E & Contrastive 2 & 33.92 & - & 2 & - \\
        \midrule
        \multirow{4}{*}{est} 
        & OFF-E2E & Primary & \textbf{38.07} & - & 29.8 &  - \\
        & HF-Cascaded & Contrastive 1 & 38.00 & - & \textbf{30.2} & - \\
        & HF-E2E-ASR$_{\text{init}}$ & Contrastive 2 & 36.97  & - & 29.6 & - \\
        & HF-E2E & \multicolumn{1}{c}{-} & 36.89 & - & - & - \\
        \midrule
        \multirow{5}{*}{mlt} 
        & OFF-E2E & Primary & \textbf{57.92} & - & \textbf{67.1} & \textbf{47.87}\textsuperscript{\ddag} \\
        & HF-E2E & Contrastive 1 & 57.65 & - & 64.21 & 48.53 \\
        & HF-E2E-ASR$_{\text{init}}$ & Contrastive 2 & 57.57 & - & 63.23 & 48.65 \\
        & HF-MLT & \multicolumn{1}{c}{-} & 57.46 & - & - & -\\
        & HF-Cascaded & \multicolumn{1}{c}{-} & 57.04 & - & - & - \\
        \midrule
        \multirow{3}{*}{mar} & HF-E2E & Primary & \textbf{44.84} & \textbf{53.80} & 43.4 & - \\
        & HF-E2E-ASR$_{\text{init}}$ & Contrastive 1 & 44.72 & 53.77 & \textbf{44.3} & - \\
        & OFF-E2E & Contrastive 2 & 42.52 & 51.34 & 41.5 & - \\
        \midrule
        \multirow{5}{*}{que} 
        & HF-E2E-ASR$_{\text{init}}$-MT$_{\text{init}}$ & Primary & \textbf{13.37} & - & 12.7 & - \\
        & HF-MLT-ASR$_{\text{init}}$ & Contrastive 1 & 13.03 & - & 12.9 & - \\
        & HF-E2E-ASR$_{\text{init}}$ & Contrastive 2 & 13.00 & - & \textbf{13.0}  & -\\
        & HF-Cascaded & \multicolumn{1}{c}{-} & 13.15 & - & - & - \\
        & HF-E2E & \multicolumn{1}{c}{-} & 12.32 & - & - & - \\
    \bottomrule
    \end{tabular}
    \caption{ST results for all languages.
    \textbf{HF-*} means the model is trained with HuggingFace toolkit, while \textbf{OFF-*} refers to the official codebase.
    \textbf{Eval} refers to the official evaluation result.
    \textsuperscript{*}For gle, the results are obtained without using the synthetic data \cite{moslem2024leveraging}. 
    \textsuperscript{\dag}For aeb, Eval 1 refers to LDC20022E02 and Eval 2 refers to LDC2023E09.
    \textsuperscript{\ddag}For mlt, Eval 1 refers to CV and Eval 2 refers to Masri.
    }
    \label{tab:st_res}
\vspace{-10pt}
\end{table*}

Internal evaluation and official evaluation~\cite{abdulmumin-etal-2025-findings} results are presented in Table~\ref{tab:st_res}.
The \textbf{HF} prefix indicates models fine-tuned using the HuggingFace toolkit, while the \textbf{OFF} prefix refers to models fine-tuned with the official codebase.
For a fair comparison, we compare results obtained from the same codebase.
\textbf{E2E} refers to E2E ST fine-tuning (Section~\ref{sec:e2e}), \textbf{Cascaded} refers to the cascaded ST system (Section~\ref{sec:cascade}), and \textbf{MLT} denotes multi-task fine-tuning (Section \ref{sec:mlt}).
\textbf{ASR$_{\text{init}}$} and \textbf{MT$_{\text{init}}$} indicate that the speech encoder and the text decoder are initialized with the fine-tuned ASR encoder and MT decoder, respectively (Section \ref{sec:pretrain}).
For \verb|gle|, results are reported without using the synthetic ST dataset \cite{moslem2024leveraging}, as we observed a performance drop when including it.
Additionally, we report the zero-shot performance of \verb|SeamlessM4T-v2-Large| in Table~\ref{tab:0shot_st}.

\textbf{The official codebase yields stronger performance.}
In Table \ref{tab:st_res}, E2E ST fine-tuning using the official codebase performs strongest in 5 out of 9 languages.
It is unexpected that the official codebase (OFF-E2E) outperforms the HuggingFace codebase (HF-E2E) in all languages except for \verb|bem| and \verb|mar|.
We discuss the discrepancies in Appendix \ref{app:codebase}.

\textbf{E2E ST fine-tuning produces strong models in general.}
Compared to the zero-shot \verb|SeamlessM4T-v2-Large| performance in Table \ref{tab:0shot_st}, E2E ST fine-tuning leads to substantial improvements.
For \verb|aeb|, \verb|bem|, \verb|fon|, and \verb|que| whose zero-shot BLEU scores are close to 0, E2E fine-tuning improves by about +20, +30, +40, and +10 BLEU, respectively.
For languages where \verb|SeamlessM4T-v2-Large| has good performance already, E2E fine-tuning yields improvements of at least +10 BLEU, except for \verb|gle|, which has a modest gain of +1 BLEU.
Overall, E2E ST fine-tuning (including both OFF-E2E and HF-E2E) achieves the best performance in 6 out of the 9 languages.
Notably, for \verb|bem|, the E2E ST result even surpasses the MT result by about 2 BLEU.

\textbf{E2E ST fine-tuning performs best for languages with ASR support.}
Next, we compare different fine-tuning strategies.
For a fair comparison, we compare results obtained by the HF codebase.
HF-E2E performs best in \verb|gle|, \verb|mlt| and \verb|mar|, exactly the languages that \verb|SeamlessM4T-v2-Large| provides ASR support.
Having been trained on large amounts of ASR data, \verb|SeamlessM4T-v2-Large| already has a strong capability to extract semantics from speech in those languages.
Further fine-tuning on our own small ASR datasets may just hurt the model's generalization capability.
However, ASR encoder initialization has only a minor negative effect, with a performance drop less than 1 BLEU.

\textbf{In-domain pre-training improves performance for languages without ASR support.}
For languages that \verb|SeamlessM4T-v2-Large| does not support ASR for, fine-tuning on in-domain ASR datasets improve the ST performance.
Specifically, for \verb|bho| and \verb|aeb|, ASR training improves performance by about +5 and +3 BLEU, respectively.
Smaller gains of about +1 BLEU are observed for \verb|bem| and \verb|que|, while the remaining languages see improvements of less than 1 BLEU.
In contrast, text decoder initialization is less effective.
It provides a slight improvement for \verb|que| but hurts \verb|aeb| performance.

\textbf{Multi-task training is beneficial when MT performance is strong.}
We explored multi-task training for \verb|aeb|, \verb|mlt|, and \verb|que|, languages for which fine-tuned MT models outperform E2E ST models.
The gaps are approximately 8, 5, and 2 BLEU, respectively.
Multi-task improves \verb|aeb| performance by 2 BLEU and improves \verb|que| by about 0.7 BLEU.
However, there is no improvement for \verb|mlt|.

\textbf{Cascaded systems are competitive but generally underperform E2E ST fine-tuning.}
We evaluate cascaded systems for \verb|aeb|, \verb|bem|, \verb|est|, \verb|mlt|, and \verb|que|.
Among these, the cascaded system only outperforms E2E ST fine-tuning in \verb|est|, with a modest gain of +1 BLEU.
For \verb|bem| and \verb|mlt|, cascaded systems even underperform direct E2E ST fine-tuning.
For \verb|aeb| and \verb|que|, although cascaded systems are better than direct E2E ST fine-tuning, they fall short compared to ST models initialized with in-domain pre-trained components.

\section{Conclusion and Future Work}
In this paper, we describe GMU systems for the IWSLT 2025 Low-resource ST shared task.
We focus on fine-tuning the \verb|SeamlessM4T-v2-Large| model and explore four fine-tuning strategies.
We find that E2E ST fine-tuning performs best on languages with ASR support.
For languages without ASR support, we can fine-tune the model on in-domain ASR datasets first and then initialize the ST encoder with the ASR encoder, which significantly improves performance.
Multi-task training and cascaded systems are not as good as E2E fine-tuning in general.
We hypothesize that it is because \verb|SeamlessM4T-v2-Large| is strong enough on ST, and the fine-tuned MT performance is not strong enough to provide useful additional performance gains.

For future work, we could explore better pre-training methods to mitigate the gap between the speech encoder and the text decoder \cite{le2023pre}.
We could also explore the use of speech large language models, as large language models have recently achieved success in MT tasks \cite{kocmi-etal-2024-findings}.

\section*{Acknowledgements}
We are thankful to the Shared Task organizers and the anonymous reviewers for their valuable feedback. This work is supported by the National Science Foundation under awards IIS-2327143 and CIRC-2346334.

\bibliography{anthology, custom}

\appendix

\section{Discrepancies between codebases}
\label{app:codebase}

There are three discrepancies between our HuggingFace codebase and the official codebase.

\textbf{Loss on the target language code.}
During training, the target sequence is formatted as $\left[</\text{s}>, <\text{lang}>, \text{token}_1, \ldots, \text{token}_n, </\text{s}>\right]$, where $</\text{s}>$ is both the start-of-sentence and end-of-sentence token, and $<\text{lang}>$ denotes the language code.
The text decoder takes as input $\left[</\text{s}>, <\text{lang}>, \text{token}_1, \ldots, \text{token}_n\right]$.
The losses described in Section \ref{sec:methods} are computed using $\left[<\text{lang}>, \text{token}_1, \ldots, \text{token}_n, </\text{s}>\right]$ as the label.
The official codebase ignores the loss on the first label, i.e., $<\text{lang}>$.
However, we still include this loss, because we use the same codebase for ASR and we want to train the language code embedding for newly added languages like \verb|<aeb>|.

\textbf{Parameter sharing of word embeddings.}
There are three word embeddings in a SeamlessM4T model: a text encoder embedding, a text decoder input embedding, and a text decoder output embedding (also termed \texttt{lm\_head}).
These three embeddings are intended to share the same weight matrix.
However, in the official codebase, the \texttt{lm\_head} is accidentally untied from the other two embeddings during model initialization, resulting in additional 262M trainable parameters.
In contrast, the HuggingFace codebase still ties all three embeddings.

\textbf{Dropout modules.}
There are a few dropout modules in the HuggingFace model that differ from the official model.
\begin{enumerate}
    \item \verb|ffn_dropout| in the decoder layers:
    The HuggingFace model uses $p=0.0$, whereas the official model uses $p=0.1$.
    \item \verb|dropout| in the \verb|self_attn| module of the adapter layer:
    The HuggingFace model uses $p=0.0$, while the official model uses $p=0.1$.
    \item \verb|intermediate_dropout| in the \verb|ffn| module of the adapter layer:
    The HuggingFace model uses $p=0.1$, while the official model uses $p=0.0$.
    \item There is a dropout module with $p=0.1$ applied to the text decoder input word embedding in the official model, but it is missing in the HuggingFace model.
\end{enumerate}
We are able to fix the first 3 dropout modules easily.
However, adding a missing dropout module for the last one would require some more efforts, so we leave it unresolved for now.

We also present experiment results on \verb|aeb| after addressing these discrepancies.
As Table~\ref{tab:st_fix} shows, the HuggingFace model achieves performance comparable as the official model for E2E ST after resolving all three discrepancies (+\verb|lm_head|+dropout+lang).
Addressing only a single or two of the discrepancies does not have a significant effect.

\begin{table}[h]
    \centering
    \begin{tabular}{lcc}
    \toprule
       \multirow{2}{*}{System}  & Dev & Public Test  \\
         & BLEU & BLEU \\
    \midrule
    OFF-E2E & 23.76 & 19.67 \\
    \midrule
    HF-E2E & 22.73 & 18.35 \\
    +\verb|lm_head| & 22.88 & 19.14 \\
    \hspace{3mm} +dropout & 22.88 & 19.22 \\
    \hspace{3mm} +lang & 22.36 & 18.86 \\
    \hspace{8mm} +dropout & \textbf{23.50} & \textbf{20.12} \\
    +dropout & 22.58 & 19.52 \\
    \bottomrule
    \end{tabular}
    \caption{ST results for aeb.
    \textbf{+lm\_head} means the lm\_head is untied from word embeddings.
    \textbf{+dropout} means we use the same drop modules as in the official model.
    \textbf{+lang} means we do not compute loss on the target language code.
    Combining all three changes yields comparable performance as the official codebase.
    }
    \label{tab:st_fix}
\end{table}

\section{Ablation study of using additional datasets}
\label{app:ablate}
In this section, we present results when using different amounts of ST training data for \verb|gle| and \verb|que|.

\textbf{gle-eng}.
There are approximately 7 hours of official 2-way ST data and about 200 hours of synthetic 3-way ST data \cite{moslem2024leveraging} available for \verb|gle|.
We attempted to incorporate the synthetic 3-way ST data  into E2E ST fine-tuning.
However, it did not help as shown in Table \ref{tab:gle_ablate}.
When training on the official ST dataset only, the dev set performance is 29.63 BLEU.
In contrast, training on both the official and synthetic data results in a performance drop of 1 BLEU.

\begin{table}[h]
    \centering
    \begin{tabular}{lcc}
        \toprule
        \multirow{2}{*}{Datasets} & \multicolumn{1}{c}{Dev} & \multicolumn{1}{c}{Public Test} \\
         & BLEU & BLEU \\
         \midrule
         IWSLT2025 & \textbf{29.63} & 51.91 \\
         \hspace{3mm} +\citet{moslem2024leveraging} & 28.69 & 51.46 \\
         \bottomrule
    \end{tabular}
    \caption{gle-eng results on the IWSLT2025 dev set.
    All models are trained using the official codebase.}
    \label{tab:gle_ablate}
\end{table}

\textbf{que-spa}.
There are only 1.67 hours of official 3-way ST data for \verb|que|.
Additional resources include approximately 8 hours of synthetic 3-way data~\cite{zevallos2022huqariq}, about 12k lines of MT data~\cite{ortega2020neural}, and about 48 hours of ASR data~\cite{cardenas2018siminchik}.
We also created a synthetic \verb|que|-\verb|spa| MT dataset using the NLLB~\cite{nllb2022} \verb|que|-\verb|eng| alignments, resulting in approximately 34k lines of bitext.
Details have been described in Section \ref{sec:dataset}.

The ASR, MT, and E2E ST results are presented in Table \ref{tab:que_asr_ablate}, Table \ref{tab:que_mt_ablate}, Table \ref{tab:que_st_ablate}, respectively, which show that incorporating all available datasets improve the performance across all three tasks.
For ASR, using additional data reduces CER by 3.65 and WER by 12.98 in absolute value.
For MT, incorporating \citet{zevallos2022huqariq} and \citet{ortega2020neural} substantially improves the performance by 8.5 BLEU.
Although the synthetic NLLB dataset is the largest, adding it only yields a marginal further improvement of 0.91 BLEU.
For ST, adding the synthetic dataset significantly improves E2E ST by 8.59 BLEU.
While the gains are smaller for E2E-ASR$_{\text{init}}$ and E2E-ASR$_{\text{init}}$-MT$_{\text{init}}$, the additional data still improves the performance by 3.16 and 2.95 BLEU, respectively.
The ASR and MT models used for initialization are the best ones from Table \ref{tab:que_asr_ablate} and Table \ref{tab:que_mt_ablate}, respectively.

\begin{table}[h]
    \centering
    \begin{tabular}{lcc}
        \toprule
        \multirow{2}{*}{Datasets} & \multicolumn{2}{c}{Dev} \\
         \cmidrule{2-3}
         & CER & WER \\
         \midrule
         IWSLT2025 & 19.19 & 50.78 \\
         \hspace{3mm} +\citet{zevallos2022huqariq} & 16.97 & 41.14 \\
         \hspace{8mm} +\citet{cardenas2018siminchik} & {\textbf{15.54}} & {\textbf{37.80}} \\
         \bottomrule
    \end{tabular}
    \caption{ASR results on the ASR split of the official 3-way ST dev set.
    }
    \label{tab:que_asr_ablate}
\end{table}

\begin{table}[h]
    \centering
    \begin{tabular}{lc}
        \toprule
        \multirow{2}{*}{Datasets} & Dev \\
         & BLEU \\
         \midrule
         IWSLT2025 & 5.88 \\
         \hspace{3mm} +\citet{zevallos2022huqariq} & \multirow{2}{*}{14.38} \\
         \hspace{3mm} +\citet{ortega2020neural} & \\
         \hspace{8mm} +\citet{nllb2022} & {\textbf{15.29}} \\
        \bottomrule
    \end{tabular}
    \caption{MT results on the MT split of the official 3-way ST dev set.}
    \label{tab:que_mt_ablate}
\end{table}

\begin{table}[h]
    \small
    \centering
    \begin{tabular}{lcc}
    \toprule
        \multirow{2}{*}{Datasets} & \multirow{2}{*}{System} & Dev \\
         & & BLEU \\
         \midrule
         \multirow{3}{*}{IWSLT2025} & E2E & 3.73 \\
         & E2E-ASR$_{\text{init}}$ & 9.84 \\
         & E2E-ASR$_{\text{init}}$-MT$_{\text{init}}$ & 10.42 \\
         \midrule
         \multirow{3}{*}{+\citet{zevallos2022huqariq}} & E2E & 12.32 \\
         & E2E-ASR$_{\text{init}}$ & 13.00 \\
         & E2E-ASR$_{\text{init}}$-MT$_{\text{init}}$ & \textbf{13.37} \\
         \bottomrule
    \end{tabular}
    \caption{ST results on the ST split of the official 3-way ST dev set.
    Models are trained using the HuggingFace codebase.
    The ASR and MT models are the best ones trained on all available ASR and MT datasets, respectively.
    }
    \label{tab:que_st_ablate}
\end{table}

\end{document}